# Dealing with Metonymic Readings of Named Entities

**Thierry Poibeau (thierry.poibeau@lipn.univ-paris13.fr)**
Laboratoire d'Informatique de Paris-Nord, Université Paris 13 and UMR CNRS 7030
99, avenue Jean-Baptiste Clément – 93430 Villetaneuse – France

**Abstract**

The aim of this paper is to propose a method for tagging named entities (NE), using natural language processing techniques. Beyond their literal meaning, named entities are frequently subject to metonymy. We show the limits of current NE type hierarchies and detail a new proposal aiming at dynamically capturing the semantics of entities in context. This model can analyze complex linguistic phenomena like metonymy, which are known to be difficult for natural language processing but crucial for most applications. We present an implementation and some test using the French ESTER corpus and give significant results.

## Introduction

Categorization is a key question in science and philosophy at least since Aristotle. Many research efforts have been made on this issue in linguistics since text understanding and more generally, reasoning or inferring largely require a precise identification of objects referred to in discourse. Lexical semantics has attracted the major part of research related to these issues in linguistics in the last few years. What is the meaning of an expression? How does it change in context? These are still open questions.

Many research projects have addressed the issue of proper name identification in newspaper texts, especially the Message Understanding Conferences (MUC-6, 1995). In these conferences, the first task to achieve is to identify named entities (NE), i.e. proper names, temporal and numerical expressions. This task is generally accomplished according to a pre-defined hierarchy of entity categories. The categorization process relies on the assumption that NEs directly refer to external objects and can thus be easily categorized. In this paper, we show that this assumption is an over-simplification of the problem: many entities are ambiguous and inter-annotator agreement is dramatically low for some categories.

We assume that even if NE tagging achieves good performances (over .90 rate of combined precision and recall is frequent on journalistic corpora), NEs are intrinsically ambiguous and cause numerous categorization problems. We propose a new dynamic representation framework in which it is possible to specify the meaning of a NE from its context.

In the paper, we report previous work on NE tagging. We then show different cases of polysemous entities in context and some considerations about their referential status. We detail our knowledge representation framework, allowing to dynamically compute the semantics of NE sequences from their immediate context. Lastly, we present an implementation and some experiments using the French ESTER corpus and showing significant improvements.

## Names, categorization and reference

There is a kind of consensus on the fact that categorization and reference of linguistic expressions are related to discrete-continuous space interplay. Categorization is the ability to select parts of the environment and classify them as instances of concepts. The process of attention is then the ability to specifically focus on a part of the observation space that is relevant in a given context (Cruse and Croft, 2004). Selected parts of the observation space is said to be *salient*.

Two important linguistic phenomena are based on a shift in the meaning profile of a word: the highlighting of its different facets and the phenomenon of metonymy (Nunberg, 1995) (Fass, 1997). A metonymy denotes a different concept than the "literal" denotation of a word, whereas the notion of facet only means focusing on a specific aspect of a concept (different parts of the meaning space of a word or "different ways of looking at the same thing"). However, both phenomena correspond to a semantic shift in interpretation ("profile shift") that appear to be a function of salience (Cruse and Croft, 2004).

In this section, we examine different theories concerning this topic, especially the model proposed by Pustejovsky (1995). We then discuss the case of NEs and examine previous work dealing with related questions using Natural Language Processing techniques.

### Pustejovsky's Generative lexicon (1995)

Pustejovsky developed an interesting model for sense selection in context (1995). His proposal – the Generative Lexicon – is based on Davidson's logic model and a strict typed theory developed in Pustejovsky (1995) and more recently in Asher and Pustejovsky (1999). Words like *book* are called `dot object`: "dot" is a function enabling to encode two facets of a given word. A *book* is by default a physical object but some verbs like *read* or *enjoy* might activate specific features that coerce the initial type: *book* then no longer refers to a physical object but to its content (through its "telic role" encoded in a complex structured called the qualia structure). Moreover, complex operations related to the same process explain why *John enjoyed his book* is interpreted as an ellipsis and imply *reading a book*.

As we will see in the next section, the same phenomenon is observed for NEs, on an even larger scale when the source is broadcast news corpora.

The existence of dot-objects should be discussed in much more detail (see Fodor and Lepore, 1998). Dot-objects enable a thorough analysis of the above example. However, even if some kind of inheritance exists in the Generative Lexicon, dot-objects are typed in a way which tends to separate rather than to gather core word semantics. Pustejovsky gives examples such as *he bought and read this book* where *book* refers to a physical object and then to the content of this physical object in the same sentence. Pustejovsky also speculates that there is a default interpretation for a sentence like *John began the book*, which means, from his point of view, that *John began reading the book*. The verb *read* is integrated as a default value for the telic role of *book* (encoded in the qualia structure).

From a cognitive point of view as well as on a linguistic basis, it seems difficult to accept that the sequence *book* receives two different types in the same sentence, depending on the context[1]. We think that strict typing is not cognitively plausible and partly occults the meaning of the whole sentence. We think that there is a unique meaning of *book* (which means only one type) and the context only highlights some of the specificities (*ways of seeing*, which can assimilated to *facets*) of the word. More precisely:

- There is no default value for interpretation but, depending on the context, some interpretations are preferred to others, as explained by Wilks (1975).
- Reference is not always explicit. *John enjoyed the book* does not only refer to the sole act of reading nor to any implied predicate, but to a complex set of under-specified features carried by the whole sentence.
- Names (including proper names) are complex units referring to continuous meaning spaces. Specific focalisation can temporally be given depending on the context.
- This focalisation can be deduced from the context, using surface methods to compute salient features in context.

Some studies already gave some evidence for such a theory. A recent and important contribution to this problem has been given by Lapata and Lascarides (2003): they show, using a probabilistic surface model measuring co-occurrences in a large tagged corpus, that *begin a book* does not select only *read*, but also *write*, *appear in*, *publish*, *leaf through*, etc.

This phenomenon is dramatically important in real texts. It is especially crucial for NEs that should receive an appropriate type depending on the context. Text understanding and machine translation for example may require such typing.

---

[1] Copestake and Briscoe (1995) propose a model to deal with metonymy, using lexical rules implemented in the framework of a unification-based framework. This approach completely avoids the limits of Pustejovsky's approach.

**Automatic metonymy resolution**

In the 1980's, cognitive linguistics gave an interesting contribution to meaning modelling with the use of schema to explain meaning of expressions in context. However, these results are hardly applicable for computation (Langacker, 1987).

Since the 1990's, lots of systems have been developed to automatically tag named entities from text. On the one hand, some systems use a set of manually developed patterns that will be applied on the text to accurately recognize and tag (MUC-6, 1995); On the other hand, fully automatic learning-based systems use Machine Learning techniques to learn a model in order to accurately tag texts. (see the CONLL conferences proceedings[2]).

More recently, Nissim and Markert (2003) gave an important contribution to the analysis of metonymic readings of NEs. They argue that examples such as:

*Ask seat 19 whether he wants to swap*

are very rare in real texts. Most metonymies correspond to regular shift in the meaning of expressions, like:

*Pakistan had won the World Cup*
*England won the World Cup*
*Scotland lost in the semi-final*

In these examples, the national sport team is referred to by the name of the country. This kind of phenomenon appears to be rather common. For location names, Nissim and Marckert identify more than 20% of occurrences that are metonymic use. They also identify a general pattern called `place-for-people` (a place name is used to refer to people living in that country) that corresponds to more than 80% of the non-literal use of location names.

To automatically process these cases, Nissim and Markert propose a supervised machine learning algorithm, which exploits the similarity between examples of conventional metonymy. They show that syntactic head-modifier relations are a high precision feature for metonymy recognition but suffer from data sparseness. They partially overcome this problem by integrating a thesaurus and introducing simpler grammatical features, thereby preserving precision and increasing recall.

We propose to generalize the approach from Nissim and Markert to other types of entities, using a larger French corpus. Moreover, we are not interested in the performance of the resolution algorithm as such, but we propose a knowledge framework explicitly marking the focalisation derived from the context.

---

[2] The "shared task" of the 2002 and 2003 Conference on Computational Natural Language Learning (CoNLL-2002 and CoNLL-2003) was devoted to "Language-Independent Named Entity Recognition" (see `http://www.cnts.ua.ac.be/conll2002/ner/` and `http://www.cnts.ua.ac.be/conll2003/ner/`).

# NE categorization

NE categorization is mainly based on the hypothesis that entities are referential and should receive a unique semantic type corresponding to their referent. We detail in this section complex cases for NE tagging.

**Polysemous NEs**

A brief corpus analysis shows that most entities refer to several semantic classes in context. For example, dates and events are often confused:

*September 11th was a disaster for America.*

September 11[th] should be considered both as a date and an event.

It is sometimes difficult to classify an organization name as an institution, a set of individuals or a building, even if most taxonomies propose these different semantic types.

*The journalist is speaking from the UN.*
*The UN was on strike yesterday.*
*The UN celebrated its 50[th] birthday.*
*The UN will not accept such a decision.*

The same phenomenon is active for location names:

*France wants to keep the head of IMF.*

Person names are even more variable. Let's keep apart examples where a person's name corresponds in fact to a company name (*Ferrari*) or to a building (*Guggenheim*). Lots of examples show moving categorization issues: a person name sometimes refers to a specific work, an object or whatever element related to the concerned person.

*I have Marcel Proust on this rack.*
*Peter is parked on the opposite side of the street.*

Metonymy alter the referential properties of NEs in context. There are other well-known phenomena where a person's name does not make any reference to the traditional referent: in the sentence *this man is a Schwarzenegger*, one does not directly refer to *Schwarzenegger*. This figure known as *antonomasia* is relatively frequent in literature, event in scientific papers.

The most well known example of ambiguous NE is *Prix Goncourt*, introduced by Kayser (1988). Kayser distinguished seven different meanings for this phrase: with the appropriate context, it refers to the literature award, to the book that received the award, to the corresponding sum of money, to the institution, *etc*. These examples show that NEs are not so different from other ambiguous linguistic units.

**Entity type hierarchies**

Previous work on NE recognition has traditionally been performed on news texts. People try to identify 3 types of expressions:

- ENAMEX: Proper names, including names of persons, locations and organizations.
- TIMEX: Temporal expressions such as dates and time.
- NUMEX: Numerical expressions such as money and percentages.

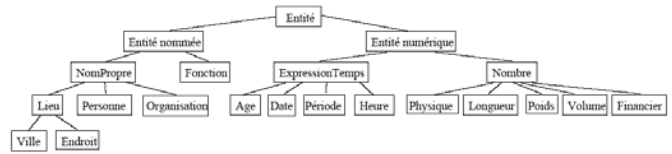

**Figure 1:** a named entity type hierarchy

Simplest hierarchies are made of a dozen of such basic types (basic types are leaves of the inverted tree) but need most of the time to be extended to cover new domains. Hierarchies of more than 200 different semantic types of entities are now common (Sekine, 2004). There is thus a need for automatic named entity recognition and disambiguation, including strategies for ambiguous items.

# Knowledge representation framework

We have shown that a fine-grained semantic categorization of NEs should not consider exclusive tags but should activate dynamic features in relation with the context of appearance of a given linguistic item (representation is inspired by the feature bundles of DATR, Evans and Gazdar, 1996).

A type hierarchy of named entities has to be defined. Proposals to refine and augment the NE hierarchy have faced problems with polysemy as shown above (Sekine, 2004). For example, what is the meaning of the *UN* in the examples given in section 3? Is it an institution, a building or a set of people?

We show that the *UN* refers to an organization, whatever its the context. We propose to introduce a focalization feature to code the salient property of the NEs in context. For example, in *The UN will not accept such a decision* the salient feature concerns the diplomatic aspect of the organization. We thus have:

```
Entity{
    Lexical_unit=ONU;
    Sem{
      Type=organization;
      Focalisation=diplomatic_org; }
}
```

The focalization feature to more specifically tag the *UN* as a diplomatic organization is activated in this context. It would be completely different in the following example:

*The news is presented this evening from the UN.*
```
Entity{
    Lexical_unit=ONU;
    Sem{
      Type=organization;
      Focalisation=localisation; }
}
```

where focalization is clearly put on the building rather than on the institution.

Focalisation seems to be stable inside a given phrase[3] but may change inside complex sentences like *John bought and read the book*).

## Towards an automatic recognition of metonymic readings of NEs

The French Evalda project organized a series of evaluation campaign concerning different areas of natural language processing. The ESTER track focused on speech enriched transcription: after transcription, the text had to be enriched with different information, including named entity tags (Gravier *et al.*, 2004) (Galliano *et al.*, 2005). We participated in his experiment since it addresses sense extension and sense coercion issues[4]. In this section, we mainly focus on the recognition of metonymic readings of NEs.

### The corpus

A corpus of about 90 hours of manually transcribed radio broadcast news was given to the participants for training purposes, 8 hours of which were identified as a development set. This acoustic corpus contained shows from four different sources, namely France Inter, France Info, Radio France International and Radio Television Marocaine. Transcribed data were recorded in 1998, 2000 and 2003. The test set consists of 10 hours of radio broadcast news shows taken from the four stations of the training corpus, plus France Culture and Radio Classique. The test set was recorded from October to December 2004. It contains 103,203 words uttered by 343 speakers. About 2.5% of this corpus correspond to advertisements and is not transcribed (for more details, please refer to Gravier *et al.*, 2004). All participants were allowed to use any data recorded prior to May 2004, whether distributed specifically for the campaign or not. In this experiment, we only used manually transcribed data.

### Description of the task: metonymy processing

The chosen NE tagset is made of 8 main categories (persons, locations, organizations, socio-political groups, amounts, time, products and facilities) and over 30 sub-categories (including categories for metonymic use of NEs). The tagset considered is therefore much more complex than the one used in the NE extraction tasks of the MUC 7 and DARPA HUB 5 programs where only 3 categories are considered (however, some previous attempts to distinguish finer-gain entity types including metonymy have been done in the framework of the NIST ACE evaluation campaign: `http://www.itl.nist.gov/iad/894.01/tests/ace/`; however, the focus of this evaluation campaign was not NE recognition) The error measure used was the slot error rate (SER).

In this experiment, we only focus on metonymic readings of NEs. This fine grained classification available from the transcribed corpus, but not officially evaluated. Our aim is to evaluate to what extent we can automatically tag named entities, according to the ESTER framework, using surface information. For example, the tagset made a difference between "natural" location (`loc`: ex. *the Alps*) and "administrative" location (`gsp`: ex. *France*). For `gsp`, 3 sub-categories were distinguished, which correspond to three different metonymic readings. The system had to make a difference between *France* as a group of people (`gsp.pers`: ex. *…les habitants du Nord de la France… – inhabitants from the west part of France*), as an organisation (`gsp.org`: ex. *..la France a signé un accord... – ...France signed an agreement...*) and as a geopolitical unit (`gsp.loc`: ex. *…ils se sont retrouvés en France… – …they met up again in France*).

The transcribed corpus contains these distinctions and a detailed guideline was produced to help people tag metonymic readings (Le Meur *et al.* 2004). However, for cases such as *France* (cf. the above example), inter-annotator agreement seems to be rather low. Even if scores over 97% are obtained on the main categories, scores can decrease down to 70% for some of the sub-categories[5]. Concerning the word *France*, it seems very difficult to make a difference between `gsp.pers` and `gsp.org` since organizations are composed of persons. Both tags appear in similar contexts.

### Features

We tried to have a theory-neutral position to automatically tag sub-categories. We had access to different kinds of information directly obtained from basic tools and resources applied on the corpus. We used the Unitex environment[6] to tag the texts according to the following resources:

- The surrounding context is known to be very useful for the task. Trigger words (person's titles, locative prepositions, …) and task-specific word lists (e.g. gazetteers) are provided by means of large dictionaries.
- Morphological analysis is done using DELA dictionaries from LADL. These dictionaries provide large coverage dictionary for French and other languages. Morphological information includes part-of-speech tags and other information like number, gender, etc.
- Chunk parsing is also done using Unitex. Surface analysis is done using basic patterns implemented through finite state automata. Chunk parsing identifies major phrases without solving attachment problems.

---

[3] Except for noun phrases such as a *heavy book*. Some authors claim that only one meaning is accessible at once (Copestake and Briscoe, 1992), which is not clear in such examples.

[4] The ESTER resources (training corpus, reference corpus the annotation guidelines and the automatic scorer) are available through a package distributed by ELRA.

[5] We asked 3 students in linguistics to tag 100 examples of ambiguous NEs (metonymic and non metonymic readings). They were provided the corpus annotation guideline.

[6] `http://www-igm.univ-mlv.fr/~unitex/`

We used the VOLEM database[7] encoding French verb semantics and alternation (Fernandez et al., 2002).

- Semantic tagging is done using various existing resources: we especially used the Larousse dictionary that provides sets of synonyms for French. Below is the example of a cluster obtained from different resources (verbs directly related to *dire – to say*): *Articuler, dire, énoncer, proférer, prononcer, ânonner, débiter, déclamer, psalmodier, réciter, claironner, clamer, crier*… If the word is ambiguous, all possible tags are used (no disambiguation).

**Algorithm**

We induce from the training corpus sets of specific features to tag metonymic readings of named entities. Characteristic units or "specificities" are elements (forms or segments) that are abnormally frequent or abnormally rare in a corpus compared to another one (Lebart *et al.*, 1997). This technique can be extended to compute the specificities of complex features, and not only of lexical items.

Probability levels (Lebart *et al.*, 1997) are used to select these characteristic features. The probability levels measure the significance of the differences between the relative frequency of an expression or a feature within a group (or a category) with its global relative frequency computed on the whole corpus. They are computed under the hypothesis of a random distribution of the form under consideration in the categories. The smaller are the probability levels, the more characteristic are the corresponding forms.

Finally, the process only keeps more specific sets of features to cover positive examples. This process is roughly similar to the one proposed by Lapata and Lascarides (2003) for the study of metonymic verbs: We compute, for each feature, its discriminative power[8] (probability to get a non literal interpretation when the feature is active in a context window around the NE).

**Results**

The official score obtained for all the ESTER categories was 76.49 F-measure (harmonic mean of precision and recall on the overall test corpus). Results are comparable to the state-of-the-art for classical categories (person's names, dates, …) and lower for difficult categories (such as *artefact*). However, this score is not interesting as such since the ESTER evaluation did not take into account the score for NE sub-categories.

We then made an intensive evaluation on metonymic readings of NE. We chose the gsp category (*France*), whose sub-types are known to be among the most difficult ones in the ESTER evaluation. Below is the obtained result (P: precision; R: recall; the baseline is obtained when all gsp are tagged as gsp.loc):

---

[7] http://www.irit.fr/recherches/ILPL. This resource has been mainly developed by P. Saint-Dizier and A. Mari for French.
[8] For a detailed discussion about the model, please refer to Lapata and Lascarides (2003: 260-270).

|          | #ref | P   | R   |
|----------|------|-----|-----|
| Gsp.loc  | 1486 | .84 | .82 |
| Gsp.pers | 7    | .01 | .29 |
| Gsp.org  | 385  | .68 | .52 |
| **baseline** | | | |
| Gsp.loc  | 1486 | .64 | .82 |
| Gsp.pers | 0    | .0  | .0  |
| Gsp.org  | 0    | .0  | .0  |

**Table 1:** automatic named entity recognition, results for the ambiguous gsp category

Results are especially bad concerning metonymic uses; they are also rather low concerning recall for gsp.loc. A manual verification of the results showed that 1) the gsp.pers category is too scarce to infer any valuable rule; 2) gsp.pers and gsp.org mainly occur in the same contexts (for example as the subject of a verb that normally requires a human subject: *il exhorte l'Amérique à y croire…– …he urges America to believe in that…* where *America* is tagged gsp.pers). This last distinction between gsp.pers and gsp.org seems to be rather subjective, since persons lead organizations (lots of gsp.org are tagged as gsp.pers by our system). The distinction would require more than surface knowledge.

We made the same evaluation but only distinguished two main categories (gsp.loc and gsp.hum; the latter one is the union of gsp.org and gsp.pers and was not part of the original ESTER guideline). We obtained the following results:

|         | #ref | P   | R   |
|---------|------|-----|-----|
| Gsp.loc | 1486 | .84 | .82 |
| Gsp.hum | 392  | .63 | .64 |

**Table 2:** automatic named entity recognition, results for the gsp category (pers and org are merged)

These results are satisfactory for a complex category including metonymic readings of NEs: they are correct for manual transcription of broadcast news. They show that the distinction between organisations and humans cannot be captured by surface form analysis.

Part of non literal readings is 20%, which is comparable to the results from Nissim and Markert (2003). For the set of ambiguous NEs described in this paper, we obtain performance similar to those reported by Nissim and Markert, although the task is harder since the corpus is made of speech transcriptions.

**The process of constructional meaning**

A quick analysis of the rules shows the following elements for the sub-categorisation analysis:
- The presence of location names with different granularity is a discriminatory element for Gsp.loc

(for example, co-occurrence of a town with a country name → `Gsp.loc`).
- `Gsp.pers` and `Gsp.org` are frequently subject of speech verbs (dire, affirmer – to say…) or more generally verbs with a human subject. Identifying these verbs using semantic tagging is thus a key issue.
- If no verb can be found, noun phrases expressing human feelings are relevant cues for `Gsp.pers` and `Gsp.org` ("l'amitié entre la France et l'Irlande…" – …friendship between France and Ireland…).

Semantic tagging seems to be the key issue for the analysis (morphology and chunking play a minor role, but chunking could be useful in a more complex framework). From a cognitive point of view, this shows that the viewpoint on the entity is changing with the context, but not its mere category. It could be interesting to encode this process using the construction grammar framework (Goldberg, 1995): NEs are shaped by the surrounding context (co-occurrences of different features) as well as by different dimensions of language (syntax and semantics being the main contributors).

## Conclusion

Named entities are not unambiguous referential elements in discourse. Semantic categories have thus to be extended to cover the different cases of semantic NE polysemy. This proposal extends the classical type hierarchy proposed in the literature from the MUC conferences. This analysis can in turn be a basis for further processing stages, like nominal anaphora resolution (Salmon-Alt, 2001; Popescu-Belis *et al.*, 1998). The representation framework presented in the paper has been extended to code other aspects of NEs such that it would be possible to deal with complex noun phrase co-reference analysis like in *IBM… the American company*. The co-reference between the two noun phrases can only be solved if a unified and coherent linguistic model is used for all information concerning NEs. This issue could be related to Schanks' MOPS (1982), since it is the basis for higher understanding capabilities.